\pdfoutput=1

\documentclass[11pt]{article}
\usepackage{authblk}

\usepackage{emnlp2021}

\usepackage{times}
\usepackage{latexsym}

\usepackage[T1]{fontenc}

\usepackage{graphicx} 
\graphicspath{ {Figures/} }
\usepackage{tabularx}
\usepackage{booktabs}
\usepackage[utf8]{inputenc}

\usepackage{microtype}

\title{Temporal Adaptation of BERT and Performance on Downstream Document Classification: Insights from Social Media}

\author[]{\textbf{Paul Röttger}}
\author[]{\textbf{Janet B. Pierrehumbert}}
\affil[]{University of Oxford}
\affil[]{\texttt{\href{mailto:paul.rottger@oii.ox.ac.uk}{paul.rottger@oii.ox.ac.uk}}}

\begin{document}
\maketitle
\begin{abstract}
Language use differs between domains and even within a domain, language use changes over time.
For pre-trained language models like BERT, domain adaptation through continued pre-training has been shown to improve performance on in-domain downstream tasks.
In this article, we investigate whether temporal adaptation can bring additional benefits.
For this purpose, we introduce a corpus of social media comments sampled over three years.
It contains unlabelled data for adaptation and evaluation on an upstream masked language modelling task as well as labelled data for fine-tuning and evaluation on a downstream document classification task.
We find that temporality matters for both tasks: temporal adaptation improves upstream and temporal fine-tuning downstream task performance.
Time-specific models generally perform better on past than on future test sets, which matches evidence on the bursty usage of topical words.
However, adapting BERT to time and domain does not improve performance on the downstream task over only adapting to domain.
Token-level analysis shows that temporal adaptation captures event-driven changes in language use in the downstream task, but not those changes that are actually relevant to task performance.
Based on our findings, we discuss when temporal adaptation may be more effective.
\end{abstract}

\section{Introduction} \label{sec: intro}
Language use differs between domains and even within a domain, language use changes over time.
In different domains, different communities share different social experiences as well as topical interests and thus produce different language \citep{church1995poisson, blei2003latent}.
At different times, some topics are discussed more actively while others fade into the background \citep{church2000empirical, altmann2009beyond, pierrehumbert2012burstiness}.
For NLP tasks, model performance therefore depends at least in part on how training and test data align in terms of domain and temporality.
Sentiment analysis models trained on film reviews, for example, perform worse on restaurant reviews \citep{liu2019survey}.
Similarly, gender and age prediction models trained on one year's data perform increasingly worse on later years \citep{jaidka2018diachronic}.

The widespread use of pre-trained language models like BERT \citep{devlin2019bert} motivates additional considerations about data selection.
Such models are first trained \textit{upstream} on large unlabelled corpora to learn general-purpose language representations (\textit{pre-training}) before labelled task data is introduced \textit{downstream} in a separate training phase (\textit{fine-tuning}).
In this setting, the choice of unlabelled pre-training data influences downstream model performance like the choice of labelled fine-tuning data does.
In particular, we know that \textit{domain} information, i.e. \textit{where} pre-training data is sampled from, is highly relevant for downstream tasks.
Domain \textit{adaptation}, i.e. additional pre-training of an already-pre-trained model on domain data, has been shown to improve performance on a wide variety of in-domain downstream tasks \citep[e.g.][]{gururangan2020don}.
By contrast, there is little insight so far into the relevance of \textit{temporality} in pre-training, i.e. \textit{when} pre-training data is sampled from, as it relates to downstream tasks.

In this article, we work towards closing this research gap by investigating whether adapting BERT to time and domain can improve performance on a downstream document classification task relative to only adapting to domain.
Our hypothesis is that temporal adaptation can capture changes in language use such as topical shifts that are relevant to the downstream task, which time-agnostic domain adaptation cannot account for.

To enable our analysis, we introduce a benchmark corpus of English-language text comments sampled from the social media site Reddit over three years.
The corpus, which we call the Reddit Time Corpus (RTC), consists of a large set of unlabelled comments for adaptation and evaluation on an upstream masked language modelling task (MLM), and a smaller set of labelled comments for fine-tuning and evaluation on a downstream five-way document classification task, which we call Political Subreddit Prediction (PSP).

\begin{figure}[h]
\centering
\includegraphics[width=0.48\textwidth]{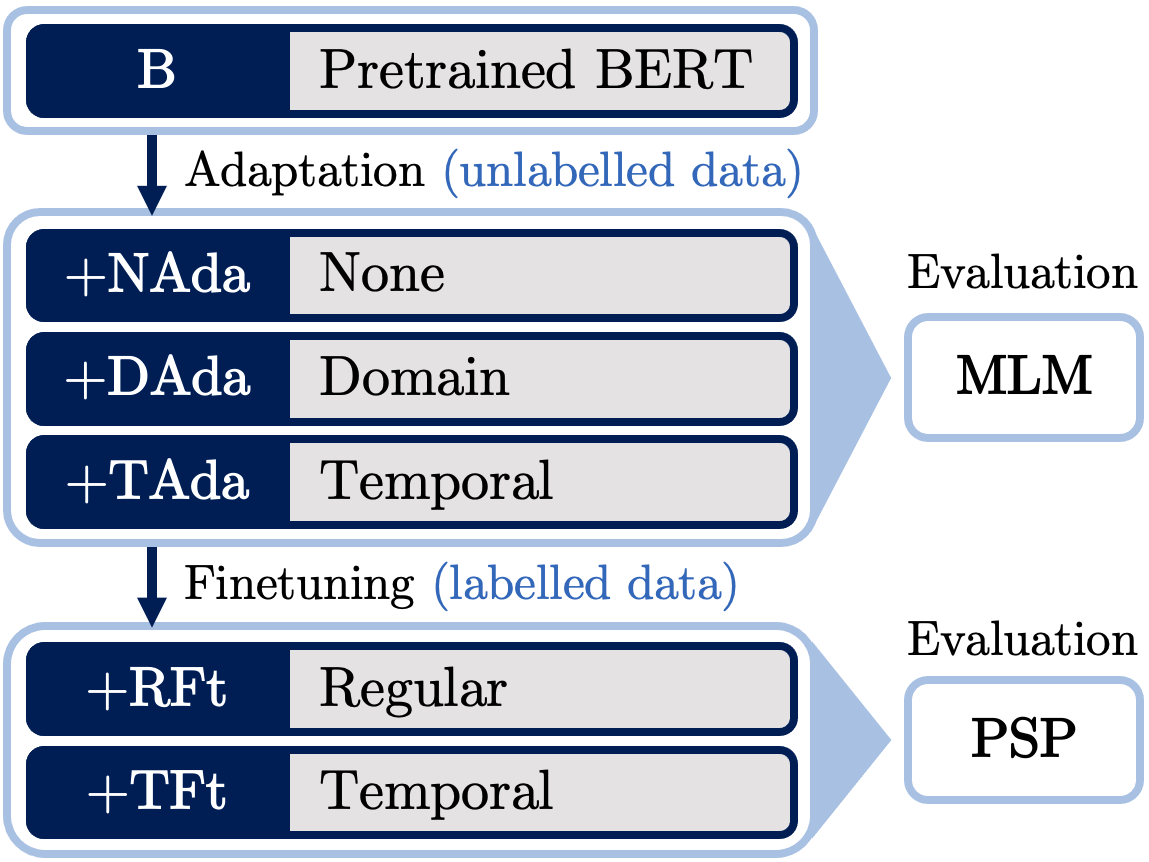}
\caption{Schematic of our experimental setup. 
BERT is first adapted in one of three ways using unlabelled data and evaluated upstream on a masked language modelling task (MLM).
Either adapted model is then fine-tuned in one of two ways using labelled data and evaluated downstream on Political Subreddit Prediction (PSP), a five-way document classification task.}
\label{fig: schematic}
\end{figure}

We use RTC and a pre-trained BERT model to conduct a series of experiments on the upstream MLM and downstream PSP task (Figure~\ref{fig: schematic}).
For MLM, we evaluate scale effects in domain adaptation (\textbf{DAda}) relative to no adaptation (\textbf{NAda}) as well as the effects of temporal adaptation (\textbf{TAda}).
For PSP, we evaluate scale effects in \textbf{DAda} in relation to regular fine-tuning (\textbf{RFt}) as well as the effects of temporal fine-tuning (\textbf{TFt}).
Lastly, we compare PSP performance across all six combinations of these adaptation and fine-tuning strategies (e.g. \textbf{TAda+TFt}).
Overall, we find that temporal information matters for both tasks.
\textbf{TAda} improves MLM performance and \textbf{TFt} improves PSP performance.
\textbf{DAda} beats \textbf{NAda} on MLM and PSP.
However, we do not find clear evidence that \textbf{TAda} outperforms \textbf{DAda} on PSP.
More granular analysis suggests that this is because the event-driven changes in language use captured by \textbf{TAda} are not discriminative, i.e. relevant, for the PSP task.\footnote{We make our code available on \href{https://github.com/paul-rottger/temporal-adaptation}{https://github.com/paul-rottger/temporal-adaptation}.}

\section{Related Work}
Previous work shows that models trained on texts from one time period perform increasingly worse on later time periods for a wide variety of tasks such as review and news article classification \citep{huang2018examining, huang2019neural}, gender and age prediction \citep{jaidka2018diachronic}, sentiment analysis \citep{lukes2018sentiment} and hate speech detection \citep{nobata2016abusive,florio2020time}.
However, such work has generally not used pre-trained models \citep[e.g.][]{jaidka2018diachronic} and even if they are used, training and evaluation focuses on labelled task data alone \citep[e.g.][]{florio2020time}.
By contrast, our analysis aims to investigate the effects of unsupervised temporal adaptation in pre-training on downstream task performance.

Within the current paradigm of using pre-trained language models, research has focused more on the domain of pre-training data than its temporality.
BERT and its variants have been pre-trained from scratch on in-domain data to improve performance on tasks such as hate speech detection \citep{tran2020habertor}, as well as tasks in scientific \citep{beltagy2019scibert}, clinical \citep{huang2019clinicalbert} and legal NLP \citep{zheng2021when}.
Further, \citet{gururangan2020don} demonstrate that domain adaptation, a second phase of pre-training on in-domain data, similarly improves performance on in-domain downstream tasks \citep[see also][]{alsentzer2019publicly,chakrabarty2019imho,lee2020biobert}.
We use their approach to domain adaptation as a baseline and extend it to temporality.

Incorporating temporal information in model pre-training has so far received little attention.
Literature on diachronic embeddings for capturing temporal semantic change \citep[e.g.][]{hamilton2016diachronic,rudolph2018dynamic,tsakalidis2020sequential} is closely related, but mostly concerned with learning representations across a known time span and investigating their dynamics.
\citet{hofmann2021dynamic} jointly model social and temporal information across time periods using a BERT model, showing that this improves performance on MLM and sentiment analysis.
However, they do not evaluate task performance across time periods.
By contrast, we adapt BERT to specific time periods with the aim of improving performance on a downstream task located in time.
More directly related to our approach, \citet{lazaridou2021pitfalls} train autoregressive, left-to-right transformer models from scratch on unlabelled data sampled up to a specific point in time and then evaluate them on a language modelling task using later data.
They find that performance degrades over time and demonstrate that dynamic evaluation \citep{krause2019dynamic}, a form of unsupervised online learning, mitigates this degradation.
By contrast, the BERT models we use learn representations through MLM and are adapted to specific time periods, which is less computationally expensive than pre-training from scratch.
Most importantly, we go beyond masked language modelling and evaluate the effects of temporal adaptation on a downstream document classification task, which is a more practically relevant use case of pre-trained language models.

\section{Experiments}

\subsection{Data: Reddit Time Corpus} \label{subsec: data}

The Reddit Time Corpus (RTC) covers three years between March 2017 and February 2020 and is split into 36 evenly-sized monthly subsets based on comment timestamps.
RTC is sampled from the Pushshift Reddit dataset published by \citet{baumgartner2020pushshift}.
We provide a data statement \citep{bender2018data} for RTC in Appendix \ref{app: data statement}.

\paragraph{Adaptation: Unlabelled News Comments}
We collect comments from \textit{r/news} and \textit{r/worldnews}, two of the most-subscribed and most active subreddits (i.e. discussion forums) on Reddit.
\textit{r/news} is primarily focused on US news content while \textit{r/worldnews} describes itself as a ``place for major news from around the world, excluding US-internal news''.
Both subreddits explicitly forbid overtly partisan posts in their community rules.
For each of 36 months in our analysis, we sample one million comments, half from each of the two subreddits, for model adaptation.
In total, we sample 36 million news comments.

\paragraph{Fine-Tuning: Labelled Politics Comments}
We collect comments from five subreddits for political discussion: \textit{r/the\_donald}, \textit{r/libertarian}, \textit{r/conservative}, \textit{r/politics} and \textit{r/chapotraphouse}.
For each of 36 months in our analysis, we sample 25,000 comments at equal proportions across these subreddits and label them by the subreddit they were posted to, to create a balanced five-way classification task with equal class distribution across months, which we call Political Subreddit Prediction (PSP).
20,000 comments are used for model fine-tuning.
5,000 comments are used for evaluation, with labels for PSP and without for MLM.
In total, we sample 0.9 million politics comments.

The subreddits we chose for PSP generally correspond to different political ideologies.
\textit{r/the\_donald} was a subreddit for supporters of then-US President Donald Trump.
\textit{r/chapotraphouse} was one of the most active leftist subreddits, which grew out of a popular podcast.
Both subreddits were shut down by Reddit in June 2020 for hosting content that promoted hate and violence.
\textit{r/conservative} and \textit{r/libertarian} are subreddits for discussing conservative and libertarian politics.
\textit{r/politics} is not explicitly ideological but its subscribers tend to be liberal-leaning \citep{marchal2020polarizing}.
We thus expect distinctions between subreddits in PSP to be at least partially predictable based on comment text for two reasons:
First, because language use differs between subreddits \citep{del2017semantic}.
Second, because distinguishing between political subreddits can be seen as a proxy for text-based ideology prediction, which is a well-established NLP task \citep[e.g.][]{conover2011predicting, iyyer2014political, kannangara2018mining, xiao2020timme}.

Since both the labelled and the unlabelled comments in RTC are sampled from the same platform, we would expect some particular degree of similarity in language use between them.
Based on Jaccard similarity of their vocabularies, comments from different politics subreddits are about as similar to each other as they are to comments from the news subreddits (Table~\ref{tab: jaccard}).
Comments from all subreddits are also more similar to each other than to paragraphs from the BooksCorpus \citep{zhu2015aligning} that was used for pre-training BERT, along with English Wikipedia content.
This motivates our use of news comments for domain adaptation.
Further, we know that topical shifts, particularly those due to exogenous events, can drive changes in language use in both news and politics comments.
For instance, Donald Trump's impeachment in December 2019 was immediately and actively discussed in news as well as politics subreddits.
This motivates our use of monthly subsets of news comments for adapting models to both domain and time.

\begin{table}[h]
\centering
\includegraphics[width=0.48\textwidth]{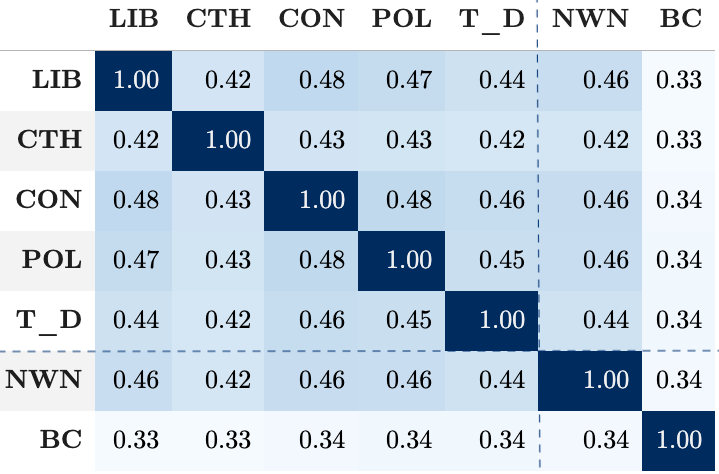}
\caption{Jaccard similarity between vocabularies for random sets of comments ($n$~=~50k) from the five political subreddits (LIB, CTH, CON, POL, T\_D) as well as the union of the two news subreddits (NWN) in RTC and a random sample of paragraphs ($n$~=~50k) from the BooksCorpus (BC) used in BERT's pre-training.}
\label{tab: jaccard}
\end{table}

\paragraph{Pre-Processing}
During sampling, we restrict RTC to English-language comments using the \texttt{langdetect} Python library.
We replace URLs and emojis with [URL] and [EMOJI] tokens, remove line breaks and collapse white space.
We remove comments posted by bots, which we identified heuristically.
We also remove comments that users have deleted from Reddit and drop duplicates within each monthly subset of the corpus.

\subsection{Model Architecture: BERT} \label{subsec: model architecture}

We use uncased BERT-base \citep{devlin2019bert} for all experiments.
For adapting to unlabelled news comments, we initialise BERT with default pre-trained weights and then continue pre-training on the MLM objective for one epoch, i.e. one pass over all additional data.
For fine-tuning on labelled politics comments, we add a linear layer with softmax output and train for three epochs.
Further details on model training and parameters as well as implementation can be found in Appendix \ref{app: parameters}.

\subsection{Upstream Task: MLM} \label{subsec: mlm}

\paragraph{Scale Effects in Domain Adaptation}
First, we evaluate the relative advantage of adapting BERT to domain using unlabelled news comments (\textbf{B+DAda}) and the extent to which this advantage scales with the amount of adaptation data.
To eliminate temporal effects, news comments for adaptation and evaluation are sampled in equal proportions across all 36 months in RTC.
As an evaluation metric, we report pseudo-perplexity, which we calculate as the exponential of the average cross-entropy loss across masked tokens (Table~\ref{tab: mlm - domain}).

\begin{table}[h]
\centering
\begin{tabularx}{0.42\textwidth}{lX}
\toprule
\textbf{Adaptation Data} & \textbf{Pseudo-Perplexity} \\
\midrule
0 (= \textbf{B+NAda}) & 19.54 \\
1 million & 8.36 \\
2 million & 7.77 \\ 
5 million & 7.10 \\ 
10 million & 6.62 \\ 
\bottomrule
\end{tabularx}
\caption{Pseudo-perplexity of \textbf{B+DAda} on overall MLM test set ($n$~=~5k unlabelled politics comments) for different amounts of adaptation data.}
\label{tab: mlm - domain}
\end{table}

MLM performance clearly benefits from adapting to domain.
Pseudo-perplexity on the test set decreases by 57.22\%, from 19.54 to 8.36, for \textbf{B+DAda} with one million news comments compared to \textbf{B+NAda}.
Performance further improves with the amount of adaptation data, although incremental improvements are diminishing.

\paragraph{Temporal Adaptation}
Second, we introduce temporality by adapting to and evaluating on comments sampled from specific months.
We adapt pre-trained BERT to one million news comments from each month in RTC, which yields 36 models (\textbf{B+TAda}).
We then evaluate each month-adapted model on each monthly test set of 5,000 politics comments, so that in total we perform 1,296 evaluations.
Pseudo-perplexity is comparable between models on the same test set but not between different test sets.
Thus, we report percentage differences in pseudo-perplexity relative to the pseudo-perplexity of a domain-adapted control model (\textbf{B+DAda} with one million news comments) on a given test set.
For readability, Table~\ref{tab: mlm - months} shows results for every fourth month.

\begin{table}[h]
\centering
\includegraphics[width=0.48\textwidth]{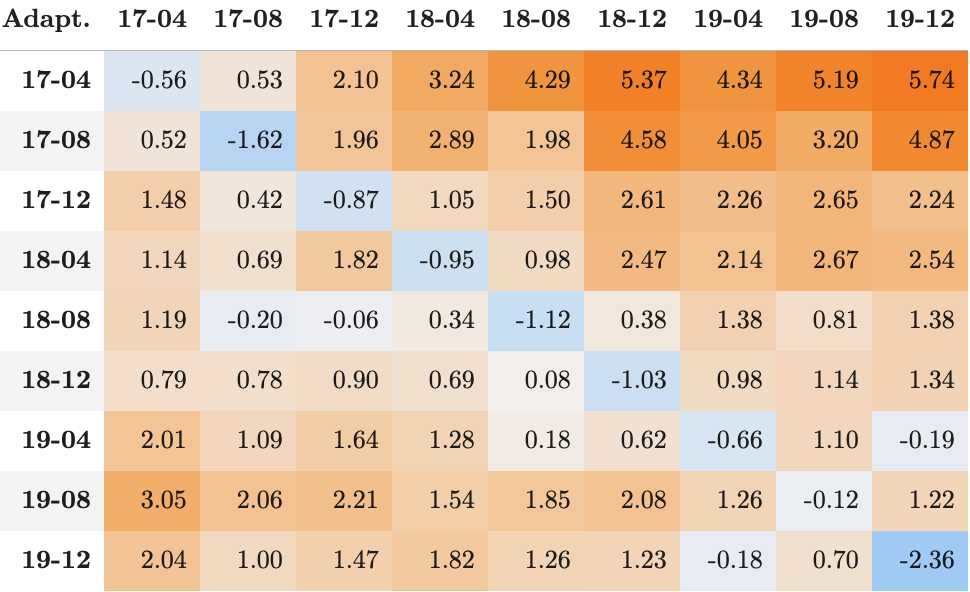}
\caption{\% difference in pseudo-perplexity of month-adapted models (\textbf{B+TAda}) relative to the control model (\textbf{B+DAda}). Rows correspond to adaptation sets ($n$~=~1m unlabelled news comments), columns to test sets ($n$~=~5k unlabelled politics comments).}
\label{tab: mlm - months}
\end{table}

For each monthly test set of politics comments, the best-performing model is the one adapted to news comments from the same month.
When adaptation month matches test month, \textbf{TAda} outperforms \textbf{DAda} by 1.03\% on average.
For other months, \textbf{TAda} generally performs worse than \textbf{DAda}.
As the temporal distance between adaptation month and test month increases, \textbf{TAda}'s performance decreases.
Lastly, relative to the month they were adapted to, models generally perform better on past than on future test data.
A one-sided Wilcoxon signed-rank test of all pairs of matching off-diagonal results (e.g. for a model adapted to 17-12, tested on 18-12 vs. a model adapted to 18-12, tested on 17-12) confirms that this finding is highly significant ($p<0.001$).

\begin{figure}[h]
\centering
\includegraphics[width=0.48\textwidth]{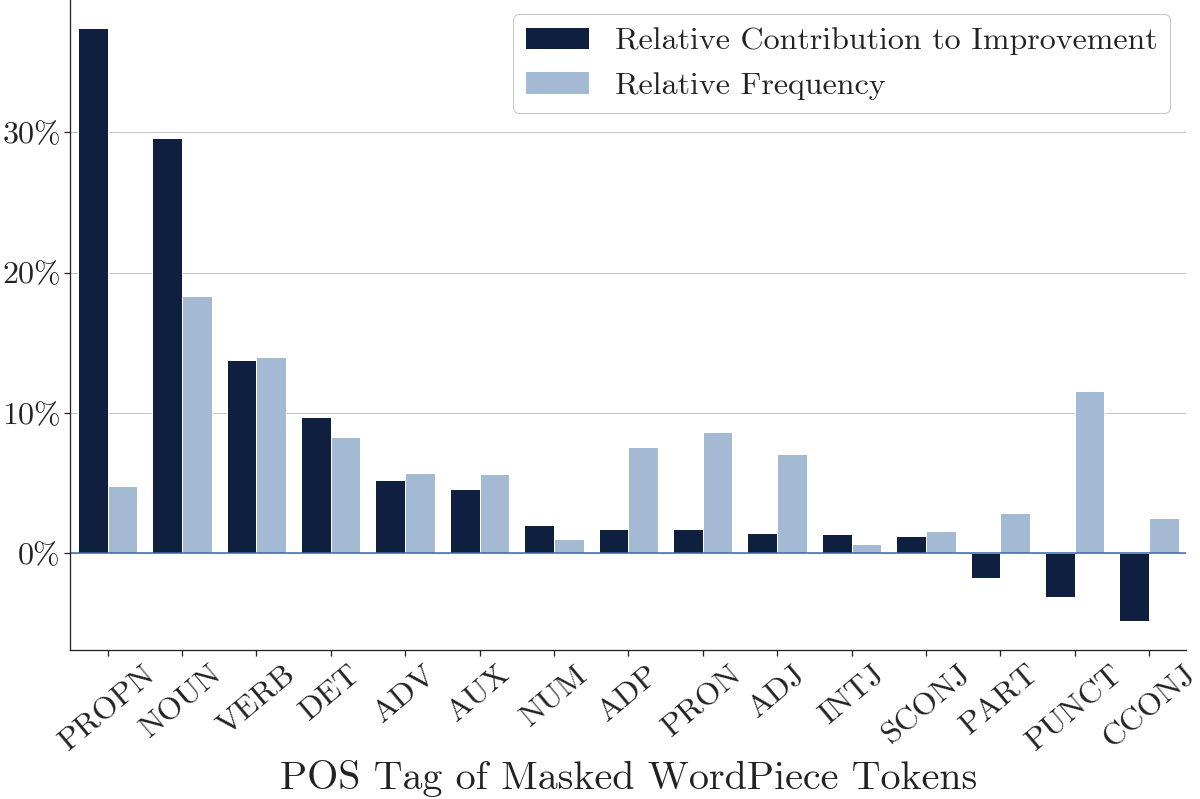}
\caption{Relative contribution to reduction in cross-entropy loss, \textbf{B+TAda} over \textbf{B+DAda} on MLM, and relative frequency of masked tokens ($n$~$\approx$~1m) by POS.}
\label{fig: mlm - pos}
\end{figure}

\paragraph{Token-Level Analysis}
To further investigate why \textbf{TAda} outperforms \textbf{DAda} on MLM when adaptation month matches test month, we analyse changes in cross-entropy loss on individual masked tokens and how they contribute to the overall performance improvement.\footnote{BERT uses a WordPiece vocabulary. Each token is an instance of a WordPiece, which may be a word or sub-word.}
Since we would expect dynamics of language use to vary between word classes, we use part-of-speech (POS) tags to structure our analysis.
Specifically, we apply the \texttt{spaCy} POS tagger to all 36 test sets, link the tags to the WordPiece tokens generated by BERT and then compare performance improvements on masked WordPiece tokens by POS tag (Figure~\ref{fig: mlm - pos}).

Masked tokens in proper and common nouns drive 67.09\% of the overall performance improvement.
The former in particular contribute disproportionately much (37.46\%) despite making up just 4.76\% of all masked tokens.
The contribution of tokens in other open-class words like verbs and adverbs roughly matches their frequency.
By contrast, tokens in closed-class words such as conjugations contribute disproportionately little.

Qualitative analysis of those tokens in proper nouns for which cross-entropy loss was reduced the most from \textbf{TAda} over \textbf{DAda} suggests that \textbf{TAda} was most effective in capturing event-driven changes in topical language use (Table \ref{tab: mlm - events}).
These changes are generally bursty.
The WordPiece "\#\#ugh" as in "Kavana\textbf{ugh}", for example, was not used as part of a proper noun in any 2017 test set.
Its use peaked when Kavanaugh was proposed for the US Supreme Court in September 2018 (107/5,000 test comments) and confirmed the month after (67/5,000).
After December 2018, it was used at most nine times per test month.

\begin{table}[h]
\small
\centering
\begin{tabularx}{0.48\textwidth}{llX}
\toprule
\textbf{Proper Noun} & \textbf{Time} & \textbf{Event}\\
\midrule
2019-\textbf{nC}ov & 20-02 & WHO Covid press conference\\
Rex Till\textbf{erson} & 18-03 & Fired by Trump\\
\textbf{Aziz} Ansari & 18-01 & Abuse allegations\\
\textbf{Kim} Foxx & 19-04 & Prosecuting Jussie Smollet case\\
Liz \textbf{Warren} & 19-11 & Presidential run\\
\textbf{Moscow} & 19-08 & Trump: "Moscow Mitch"\\
\textbf{Tide} pods & 18-02 & Meme about eating them\\
\textbf{Cv}ille & 17-08 & "Unite the Right" rally\\
Ciaram\textbf{ella} & 19-11 & Revealed as CIA whistleblower\\
Kavana\textbf{ugh} & 18-10 & Supreme Court confirmation \\
\bottomrule
\end{tabularx}
\caption{Top ten most-improved masked tokens (\textbf{bold}) in proper nouns from \textbf{TAda} over \textbf{DAda}, the test month the tokens are from and the event they correspond to.}
\label{tab: mlm - events}
\end{table}

\subsection{Downstream Task: PSP} \label{subsec: psp}

\paragraph{Scale Effects in Adaptation and Fine-tuning}
First, we evaluate relative scale effects in domain adaptation (\textbf{DAda}) and regular fine-tuning (\textbf{RFt)}.
To eliminate temporal effects, we sample news comments for adaptation as well as politics comments for fine-tuning and evaluation in equal proportions across all 36 months in RTC.
Table~\ref{tab: psp - scale} reports macro F1 on a scale from 0 to 100.
Since there are five balanced classes in PSP, random choice would yield an expected macro F1 of 20.

\begin{table}[h]
\centering
\includegraphics[width=0.48\textwidth]{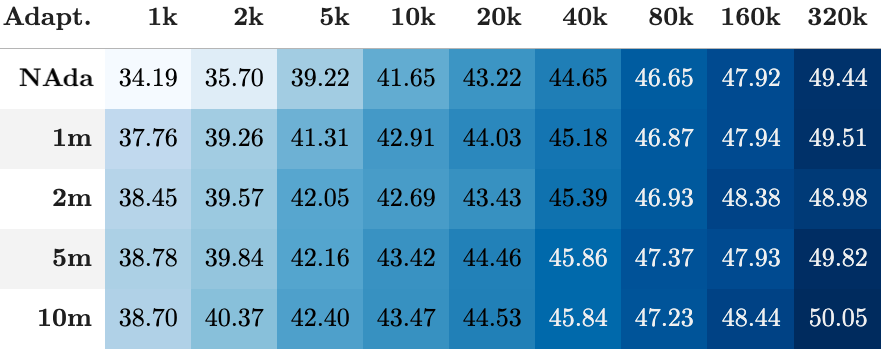}
\caption{Macro F1 of \textbf{B+DAda+RFt} on overall PSP test set ($n$~=~10k labelled politics comments). Rows correspond to different amounts of adaptation data (unlabelled news comments), columns to different amounts of fine-tuning data (labelled politics comments).}
\label{tab: psp - scale}
\end{table}

Performance monotonically increases with the amount of politics comments used for \textbf{RFt}.
Even for large amounts of fine-tuning data, there is no clear sign of a plateau.
\textbf{DAda} using news comments is relatively more effective when there is less fine-tuning data.
Its effectiveness moderately scales with the amount of adaptation data, but the biggest difference in performance is between the non-adapted model (\textbf{NAda}) and the model adapted using one million news comments, particularly for smaller amounts of fine-tuning data.

\paragraph{Temporal Fine-Tuning}
Second, we introduce temporality to PSP by fine-tuning and evaluating models on labelled politics comments from specific months (\textbf{TFt}).
We fine-tune a pre-trained, non-adapted BERT model using 20,000 politics comments from each month in RTC, which yields 36 models (\textbf{B+NAda+TFt}).
We then evaluate each month-tuned model on each monthly test set of 5,000 politics comments.
Just like pseudo-perplexity in MLM, macro F1 on PSP is comparable between models on the same test set but not between different test sets.
Therefore, we report percentage differences in macro F1 relative to the macro F1 of a control model with regular fine-tuning (\textbf{B+NAda+RFt} with 20,000 politics comments) on a given test set.
For readability, we report results only for every fourth month in Table~\ref{tab: psp - months}.

\begin{table}[h]
\centering
\includegraphics[width=0.48\textwidth]{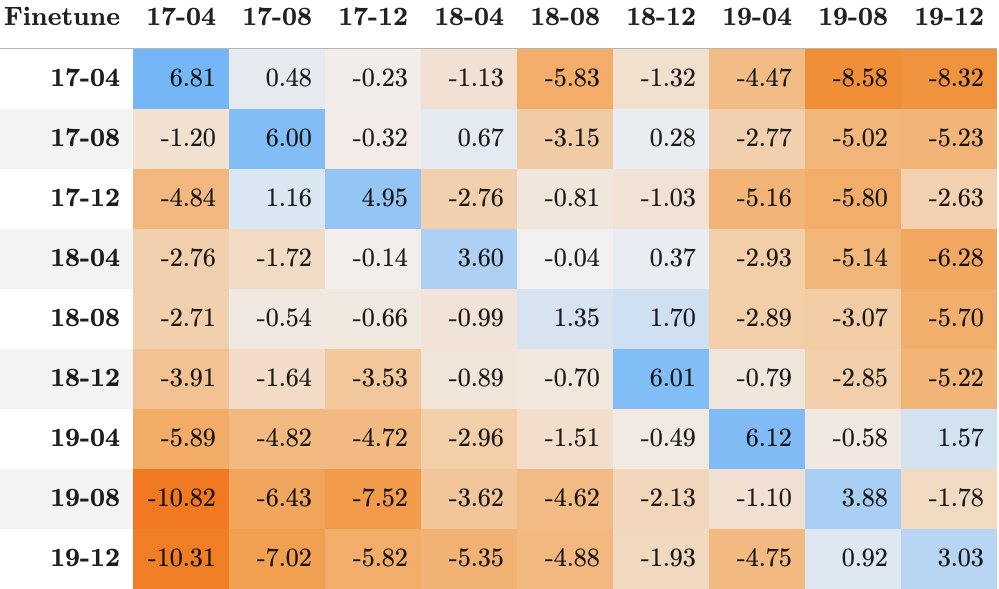}
\caption{\% difference in macro F1 of month-tuned models (\textbf{B+NAda+TFt}) relative to the control model (\textbf{B+NAda+RFt}). Rows correspond to fine-tuning sets ($n$~=~20k labelled politics comments), columns to test sets ($n$~=~5k labelled politics comments).}
\label{tab: psp - months}
\end{table}

Overall, the results for month-tuned \textbf{TFt} models on PSP resemble those for month-adapted \textbf{TAda} models on MLM (Table~\ref{tab: mlm - months}).
The best-performing model on a given test month is the one fine-tuned on politics comments from that month.
When fine-tuning month matches test month, \textbf{TFt} outperforms \textbf{RFt} by 5.09\% on average.
For other months, \textbf{TFt} generally performs worse than \textbf{RFt}, although there are some exceptions when fine-tuning and test month are not far apart.
For instance, \textbf{TFt} models on average perform 1.11\% better than the \textbf{RFt} model on the test month directly after their fine-tuning month.
As temporal distance between fine-tuning and test month grows, the performance of \textbf{TFt} models generally worsens.
Models generally perform better on past than on future test data relative to the month they were fine-tuned on.
A one-sided Wilcoxon signed-rank test of all pairs of matching off-diagonal results confirms that this finding is highly significant ($p<0.001$).

\paragraph{Adaptation and Downstream Effects}

Third, we compare PSP performance across all six combinations of adaptation (\textbf{NAda}, \textbf{DAda}, \textbf{TAda}) and fine-tuning strategies (\textbf{RFt}, \textbf{TFt}).
Our main interest is in evaluating whether \textbf{TAda} provides additional performance benefits on PSP compared to \textbf{DAda}.
As an evaluation metric, we report average macro F1 across all 36 monthly PSP test sets.
For models that incorporate temporality in adaptation (\textbf{TAda}) and/or fine-tuning (\textbf{TFt}), we consider those where adaptation and/or fine-tuning month matches the test month.
Given that we found adaptation to be more effective for smaller amounts of fine-tuning data (Table~\ref{tab: psp - scale}), we report results for fine-tuning sizes of 2,000 and 20,000 (Table~\ref{tab: psp - comparison}).

\begin{table}[h]
\centering
\begin{tabularx}{0.35\textwidth}{lcc}
\toprule
 & \textbf{2k} & \textbf{20k}\\
\midrule
\textbf{B+NAda+RFt} & 35.95 & 43.21 \\
\textbf{B+DAda+RFt} & \textbf{39.11} & \textbf{43.84} \\
\textbf{B+TAda+RFt} & 39.01 & 43.81 \\
\midrule
\textbf{B+NAda+TFt} & 37.59 & 45.41 \\
\textbf{B+DAda+TFt} & 40.19 & 46.02 \\
\textbf{B+TAda+TFt} & \textbf{40.38} & \textbf{46.12} \\
\bottomrule
\end{tabularx}
\caption{Average macro F1 across all 36 monthly PSP test sets ($n$~=~180k labelled politics comments) for the six main model configurations, split between models using \textbf{RFt} and \textbf{TFt}. Best performance is \textbf{bold}. Columns correspond to different amounts of labelled politics comments used for fine-tuning. }
\label{tab: psp - comparison}
\end{table}

Our central finding is that models adapted to time and domain (\textbf{TAda}) show no clear performance improvement over models adapted to just domain (\textbf{DAda}).
Macro F1 is marginally higher for \textbf{B+TAda+TFt} than \textbf{B+DAda+TFt}, but marginally lower for \textbf{B+TAda+RFt} than \textbf{B+DAda+RFt}.
Further, we find that \textbf{DAda} outperforms \textbf{NAda} and that \textbf{DAda} is more beneficial for models fine-tuned on less data, which matches results from Table~\ref{tab: psp - scale}.

\paragraph{MLM Improvements and PSP Performance}
Finally, we investigate why the benefits of \textbf{TAda} over \textbf{DAda} on MLM (Table~\ref{tab: mlm - months}) did not manifest in better performance on the downstream PSP task.
For this purpose, we focus on masked tokens in proper nouns, which we identified as the main driver of \textbf{TAda}'s MLM improvements (Figure~\ref{fig: mlm - pos}).

\begin{figure}[h]
\centering
\includegraphics[width=0.48\textwidth]{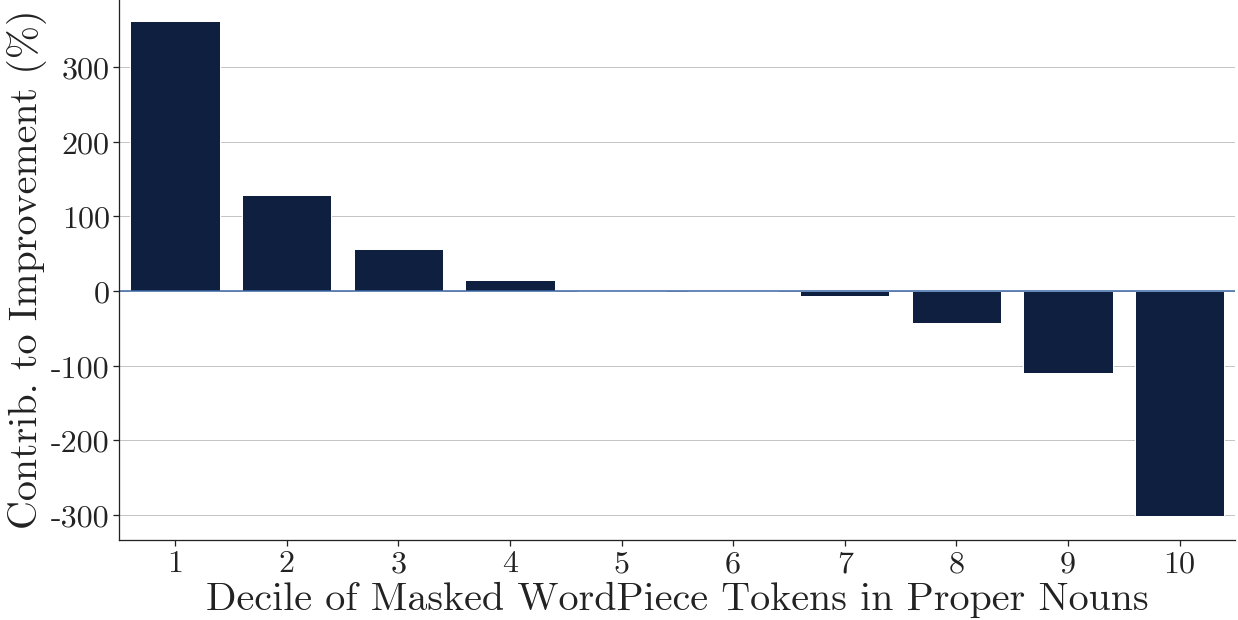}
\caption{Relative contribution to reduction in cross-entropy loss, \textbf{B+TAda} over \textbf{B+DAda} on MLM, for masked tokens in proper nouns ($n$~=~48,107) by decile.}
\label{fig: propn - improvements}
\end{figure}

Figure~\ref{fig: propn - improvements} shows that \textbf{TAda} improvements in MLM performance on masked tokens in proper nouns are overwhelmingly driven by the top 10\% most-improved tokens (e.g. "2019-\textbf{nC}ov"), which we found to closely map onto exogenous news events (Table~\ref{tab: mlm - events}).
We relate this set of most-improved tokens to PSP as follows:
For each token, we take the WordPiece it is an instance of and the PSP test set the comment with the token is from.
In that test set, we then count how many subreddits the WordPiece was used in and how many comments in each subreddit the WordPiece was used in, filtering only uses in proper nouns.
For example, one of the most-improved tokens is "Kavana\textbf{ugh}" in October 2018.
That month, the WordPiece "\#\#ugh" was used in a proper noun in 67 out of 5,000 test comments across all five politics subreddits.

\newcolumntype{C}[1]{>{\centering\let\newline\\\arraybackslash\hspace{0pt}}m{#1}}
\newcolumntype{Y}{>{\centering\arraybackslash}X}
\begin{table}[h]
\centering
\begin{tabularx}{0.47\textwidth}{cccY}
\toprule
\textbf{Subs.} & \textbf{WordPs} & \textbf{Comments} & \textbf{Avg. Freq.}\\
\midrule
\textbf{1} & 1,403 & 1,789 & 1.28 \\
\textbf{2} & 769 & 2,316 & 1.51 \\
\textbf{3} & 559 & 3,336 & 1.99 \\
\textbf{4} & 570 & 6,950 & 3.05 \\
\textbf{5} & 819 & 33,090 & 8.08 \\
\bottomrule
\end{tabularx}
\caption{Frequency measures for WordPieces corresponding to the top 10\% most-improved tokens in proper nouns by \textbf{TAda} over \textbf{DAda} for MLM ($n$~=~4,120 after deduplication).
Grouping is by the number of subreddits ($n$~=~5) that a given WordPiece was used in in a given PSP test set ($n$~=~5k).
Average frequency is calculated for the subreddits the WordPieces were used in (Avg. Freq. = Comments / WordP's / Subs.).}
\label{tab: propn - distribution}
\end{table}
Table~\ref{tab: propn - distribution} suggests that the tokens in proper nouns that drive MLM improvements of \textbf{TAda} are not relevant to the PSP task.
First, most WordPieces corresponding to these tokens are not distinctive of individual subreddits, with 2,717 WordPieces (65.95\%) used in more than one subreddit in a given test set.
The more subreddits the WordPieces are used in, the more frequent they are overall and within each subreddit they are used in.
Second, more distinctive WordPieces are much rarer.
WordPieces that are used in fewer subreddits are used much less frequently overall and used less frequently in the subreddits that they do appear in.
The 1,403 WordPieces (34.05\%) that are used in just one subreddit in a given test set are used on average in just 1.28 comments.
1,156 WordPieces are used in just one comment.

\section{Discussion}

\subsection{Results}

We find that \textbf{DAda} yields large performance improvements on upstream MLM (Table~\ref{tab: mlm - domain}) and the downstream PSP document classification task (Table~\ref{tab: psp - comparison}) when compared to \textbf{NAda}, which matches previous findings \citep{alsentzer2019publicly,chakrabarty2019imho,lee2020biobert,gururangan2020don}.
Further, \textbf{DAda} is more effective when there is little fine-tuning data (Table~\ref{tab: psp - scale}).

We also find that temporality matters for both MLM and PSP.
For upstream MLM, \textbf{TAda} outperforms \textbf{DAda} when adaptation month matches test month (Table~\ref{tab: mlm - months}).
For downstream PSP, \textbf{TFt} outperforms \textbf{RFt} when fine-tuning month matches test month (Table~\ref{tab: psp - months}).
For both tasks, model performance decreases as the temporal distance between (pre-)training and test set grows.
These findings are consistent with previous evidence for MLM \citep{lazaridou2021pitfalls} and other document classification tasks \citep[e.g.][]{huang2018examining, florio2020time}.
The results also illustrate a trade-off between temporal specificity and generalisability across time periods, which mirrors an equivalent trade-off in domain adaptation \citep{gururangan2020don}.
Further, relative to the month they were adapted to (Table~\ref{tab: mlm - months}) or fine-tuned on (Table~\ref{tab: psp - months}), models perform significantly better on past than on future test sets for MLM and PSP.
This matches evidence on the usage of topical words, which tend to occur in bursts, often triggered by an exogenous event, followed by a slower decay \citep{church2000empirical, altmann2009beyond, pierrehumbert2012burstiness}.

Despite these positive results for the individual tasks, we cannot confirm that the benefits of \textbf{TAda} over \textbf{DAda}, which are evident in MLM, transfer downstream to PSP.
\textbf{DAda} and \textbf{TAda} perform about equally well on PSP (Table~\ref{tab: psp - comparison}).
This holds for different fine-tuning strategies (\textbf{RFt} and \textbf{TFt}) and different amounts of fine-tuning data.

Several trivial explanations for this negative finding can be eliminated due to our systematic experimental approach.
First, we know that the language used in news comments is informative for PSP, since \textbf{DAda} consistently outperforms \textbf{NAda} on PSP (Tables~\ref{tab: psp - scale} and~\ref{tab: psp - comparison}).
Second, we know that discriminatory language cues for PSP change over time, since \textbf{TFt} consistently outperforms \textbf{RFt} when fine-tuning month matches test month, and since \textbf{TFt} performs increasingly worse as temporal distance between fine-tuning and test month increases (Table~\ref{tab: psp - months}).
Lastly, we know that \textbf{TAda}, which uses news comments, allows models to capture some changes in language use in politics comments, since for each monthly test set of politics comments in MLM, the best-performing model is the one adapted to news comments from that same month (Table~\ref{tab: mlm - months}).
Therefore, we can conclude that the changes in language use in politics comments that are captured by \textbf{TAda} using MLM on news comments are by and large not the changes in discriminatory language cues that are relevant to PSP.

In our token-level analysis, we find that most of \textbf{TAda}'s improvements over \textbf{DAda} on MLM (Figure~\ref{fig: mlm - pos}) are for masked tokens in nouns.
Predictions on masked tokens in proper nouns improve disproportionately much, especially for tokens that directly correspond to bursty changes in topical language use driven by exogenous news events (Table~\ref{tab: mlm - events}).
However, in relation to the PSP task, the WordPieces corresponding to these tokens generally appear non-discriminative, since most of them are used in several politics subreddits rather than just one (Table~\ref{tab: propn - distribution}).
Intuitively, many news events are not just relevant to one political ideology, although they may differ in the way they are framed \citep{card2015media, demszky2019analyzing, hofmann2021modeling}.
In March 2018, for example, when Donald Trump fired his secretary of state Rex Tillerson, an \textit{r/politics} user in the corresponding test set said they were "sympathetic" to him, while an \textit{r/the\_donald} user called him a "globalist cuck".
Since \textbf{TAda} uses comments from news subreddits, it cannot easily capture such distinctive frames.

\subsection{Promising Uses of Temporal Adaptation} \label{subsec: tada for other tasks}
Based on our findings for this particular application of \textbf{TAda}, we can formulate positive expectations about the circumstances in which \textbf{TAda} would likely be more effective.

First, we expect \textbf{TAda} to be more effective if it captured changes in language use that were more specific to individual classes in the downstream task, i.e. more discriminative.
Such changes in language use would occur when an event is relevant to just one class or when the same event is relevant to different classes at different times.
For instance, learning about a regional news event in adaptation would likely help a classifier distinguish between comments from regional news sites.

Second, we expect \textbf{TAda} to be more effective over longer time scales than the 36 months covered by RTC.
News and politics are suitable domains for our analysis because topical shifts are visible on short time scales (Figure~\ref{fig: mlm - pos}), but over decades and centuries rather than months and years, cultural shifts and linguistic drift add to shorter-term event-driven changes in language use \citep{hamilton2016cultural,hamilton2016diachronic}.
For tasks based on long-term corpora, such as the Corpus of Historical American English \citep{davies2012expanding} temporal adaptation would thus likely improve model performance.

Lastly, we may also expect \textbf{TAda} to be more effective if it used pre-training objectives that were more aligned with downstream tasks.
\citet{clark2020pretraining} argue that there is an inherent mismatch between task-agnostic pre-training that uses masked tokens and fine-tuning that does not, which recent work on discriminative pre-training tries to resolve \citep{clark2020electra}.
Future work could explore the use of such techniques for model adaptation.

Even in circumstances in which \textbf{TAda} is effective, researchers and practitioners will need to consider the performance trade-off between temporal specificity and generalisability across time periods.
For example, when deploying a hate speech detection model for content moderation, performance on newly posted content is most important, and tailoring the model to the current month is desirable even at a cost to reduced performance on past months.
However, for applications where temporality is less relevant, more heterogeneous training data sampled across months is preferable.

\section{Conclusion}
In this article, we investigated whether adapting a pre-trained BERT model to time and domain can increase its performance on a downstream document classification task compared to only adapting it to domain.
Overall, we found no clear evidence for this.
By devising a systematic experimental approach based on the novel RTC benchmark corpus, we showed that temporality is relevant for both upstream MLM and the downstream PSP document classification task.
Temporal adaptation improved MLM performance and temporal fine-tuning improved PSP performance.
Further, domain adaptation improved performance on both tasks.
Time-specific models generally performed better on past than on future test sets for both tasks, which matches evidence on the bursty usage of topical words.
However, the upstream benefits of temporal adaptation for MLM did not translate into better downstream performance on PSP compared to domain adaptation alone.
Token-level analysis showed that temporal adaptation captured event-driven changes in language use in downstream task data, but not those changes that are relevant to performance on it.
This suggests that temporal adaptation may well be effective for other tasks under circumstances we outlined, which future work could investigate.

\section*{Acknowledgments}
Paul Röttger was funded by the German Academic Scholarship Foundation.
Janet B. Pierrehumbert was supported by EPSRC Grant EP/T023333/1.
We thank Hannah Rose Kirk for helpful comments and all reviewers for their constructive feedback.

\section*{Ethics Statement}
\paragraph{Data Collection}
All data in RTC is sampled from the Pushshift Reddit dataset made publicly available by \citet{baumgartner2020pushshift}.
This dataset, in turn, was collected via Reddit's own public API in line with the site's terms of service.
Our use of this Reddit dataset was also approved by the University of Oxford's Central University Research Ethics Committee.
Labels for politics comments in RTC were created from comment metadata, so that no manual annotation was necessary.

\paragraph{Data Characteristics}
We describe the characteristics of RTC in the main body of this paper and provide additional detail in a data statement \citep{bender2018data} in Appendix~\ref{app: data statement}.
In particular, we highlight RTC's limited scope in terms of data source (Reddit) and language (English), which limits the generalisability of models trained on it.

\paragraph{Intended Use}
The intended use of temporal adaptation is as an alternative to existing strategies for continued pre-training, particularly domain adaptation.
Our article explores a specific application of temporal adaptation using monthly sets of news comments for a downstream classification task of politics comments.
Temporal adaptation could be applied to most other NLP tasks as long as pre-training and task data can be located in time, although the effectiveness of temporal adaptation may differ.
Effective temporal adaptation stands to improve diachronic model performance and thus reduce error rates in real-world applications.

\paragraph{Potential Misuse}
As with domain adaptation, temporal adaptation creates a trade-off between specificity and generalisability.
Models adapted to a particular time period and domain should not be used for other time periods and domains without careful consideration of resulting biases.

\paragraph{Environmental Impact}
Temporal adaptation is more computationally expensive than just fine-tuning using a (smaller) set of labelled task data but much less computationally expensive than pre-training from scratch on even larger unlabelled datasets.
Relative to the concerns raised around the environmental costs of the latter \citep{strubell2019energy,henderson2020towards,bender2021stochastic}, we consider the environmental costs of temporal adaptation to be relatively minor.
In practical applications, researchers could consider cumulative approaches to temporal adaptation, rather than adapting separate models for each time period, to avoid redundant computations.

\bibliography{custom}
\bibliographystyle{acl_natbib}

\clearpage
\appendix

\section{Data Statement} \label{app: data statement}

Following \citet{bender2018data}, we provide a data statement, which documents the generation and provenance of labelled and unlabelled documents in the Reddit Time Corpus (RTC).

\paragraph{A. CURATION RATIONALE}
The purpose of RTC is to enable our analysis of temporal adaptation of pre-trained language models and downstream task performance.
RTC comprises text comments that were posted to the social media site Reddit between March 2017 and February 2020.
The unlabelled portion of RTC consists of 36 million comments sampled from \textit{r/news} and \textit{r/worldnews}, two of the most active subreddits (i.e. sub-forums) on the site, which are dedicated to discussion of current news events.
The labelled portion of RTC consists of 0.9 million comments sampled in equal proportions from five subreddits for political discussion (\textit{r/the\_donald}, \textit{r/libertarian}, \textit{r/conservative}, \textit{r/politics} and \textit{r/chapotraphouse}).
Comments are labelled based on which subreddit they are from.
All data is split into 36 evenly-sized subsets based on comment timestamps.

\paragraph{B. LANGUAGE VARIETY}
RTC only contains English-language text documents, as determined by the \texttt{langdetect} Python library.
We opted for English language due to data availability.
Further, all data in RTC is sourced from Reddit.
We consider this a limitation of our analysis and suggest expansion to other languages and data sources as a priority for future research.

\paragraph{C. SPEAKER DEMOGRAPHICS}
The speakers in RTC are a sample of all Reddit users who posted a comment to one of the seven subreddits covered by RTC between March 2017 and February 2020.
In February 2020, \textit{r/worldnews} had around 23.1m subscribers, \textit{r/news} 19.9m, \textit{r/politics} 5.76m, \textit{r/the\_donald} 0.79m, \textit{r/libertarian} 0.36m, \textit{r/conservative} 0.30m and \textit{r/chapotraphouse} 0.15m.
Reddit does not make information on user demographics available but a February 2019 survey of US users indicated that roughly two-thirds were male, and that user age was skewed towards 18 to 29 years \citep{statista2019reddit}.

\paragraph{D. ANNOTATOR DEMOGRAPHICS}
We did not employ any annotators.
All labels in RTC are based on comment metadata, specifically which subreddit a given comment is from.

\paragraph{E. SPEECH SITUATION}
All comments in RTC were posted to Reddit between March 1st 2017 and February 29th 2020.
The intended audience is other subreddit users and site visitors.

\paragraph{F. TEXT CHARACTERISTICS}
All documents are individual text comments.
Pre-processing steps are described in \S\ref{subsec: data}.
For the labelled portion of RTC, we provide a label based on which of the five political subreddits in RTC they were posted to.
The class distribution is balanced in RTC overall and in each monthly subset.

\section{Model Training \& Parameters} \label{app: parameters}

\paragraph{Model Architecture}
We implemented uncased BERT-base models \citep{devlin2019bert} using the \texttt{transformers} Python library \citep{wolf2020transformers}.
Uncased BERT-base, which is trained on lower-cased English text, has 12 layers, a hidden layer size of 768, 12 attention heads and a total of 110 million parameters.
For PSP, we added a linear layer with softmax output.

\paragraph{Training Parameters}
For both MLM and PSP, we used cross-entropy loss.
As an optimiser, we used AdamW \citep{loshchilov2019decoupled} with a 5e-5 learning rate and a 0.01 weight decay.
For regularisation, we set a 10\% dropout probability.
Maximum input sequence length is 128 tokens.
For adapting to unlabelled data, we trained for one epoch, i.e. one pass over all additional data, which matches \citet{gururangan2020don}.
Training batch size was 128.
For fine-tuning on labelled data, we trained for three epochs with a batch size of 32, which corresponds to default settings recommended by \citet{devlin2019bert}.
For comparability, we used these same untuned hyperparameters across all experiments.

\paragraph{Computation}
All experiments were run between March and May 2021 using Nvidia Tesla K80 and V100 GPUs accessed through the University of Oxford's Advanced Research Computing service.
Runtime varied from experiment to experiment.
Adapting BERT to one million comments for one epoch took around three hours on a V100.
Fine-tuning BERT on 20,000 comments for three epochs took around 15 minutes.

\paragraph{Source Code}
We make all our code available at \href{https://github.com/paul-rottger/temporal-adaptation}{https://github.com/paul-rottger/temporal-adaptation}.

\end{document}